\documentclass{INTERSPEECH2023}
\usepackage{multirow}
\usepackage{graphicx}
\usepackage{tikz}
\usepackage[normalem]{ulem}
\useunder{\uline}{\ul}{}
\usepackage{tipa}
\usepackage{todonotes}
\usepackage{diagbox}

\interspeechcameraready

\newcommand{\sysname}{\textsc{LookAhead}}
\newcommand{\prnnt}{P_\text{rnnt}}
\newcommand{\lrnnt}{L_\text{rnnt}}
\newcommand{\IAM}{P_\text{iam}}
\newcommand{\blankT}{\epsilon}
\def\vx{{\mathbf x}}
\def\vy{{\mathbf{y}}}

\DeclareMathOperator*{\argmax}{argmax}

\title{Improving RNN-Transducers with Acoustic \sysname\ }
\name{Vinit S. Unni$^1$, Ashish Mittal$^{1,2}$, Preethi Jyothi$^1$, Sunita Sarawagi$^1$}
\address{
  $^1$Indian Institute of Technology Bombay, India \\
  $^2$IBM Research, India 
  }
\email{vinit@cse.iitb.ac.in, arakeshk@in.ibm.com, pjyothi@cse.iitb.ac.in, sunita@cse.iitb.ac.in}

\begin{document}

\maketitle
 
\begin{abstract}
RNN-Transducers (RNN-Ts) have gained widespread acceptance as an end-to-end model for speech to text conversion because of their high accuracy and streaming capabilities. A typical RNN-T   independently encodes the input audio and the  text context, and combines the two encodings by a thin joint network.  While this architecture provides SOTA streaming accuracy,  it also makes the model vulnerable to strong LM biasing which manifests as multi-step hallucination of text without acoustic evidence.
In this paper we propose \sysname\ that makes text representations more acoustically grounded by looking ahead into the future within the audio input.
This technique yields a significant $5\%-20\%$ relative reduction in word error rate on both in-domain and out-of-domain evaluation sets.

\end{abstract}

\section{Introduction}
RNN-Transducers (RNN-Ts)~\cite{Graves2012} are the predominant choice for end-to-end automatic speech recognition (ASR) offering both high accuracy and streaming capabilities~\cite{rao2017exploring,he2019streaming}.
They comprise a speech encoder that can process speech to generate an acoustic representation and a text encoder that is conditioned on label outputs from previous time-steps to generate a textual representation. Both the acoustic and textual representations are further combined by a simple joint network to predict the final output sequence. Apart from making the model streaming-friendly, separate speech and text modules in RNN-Ts also allow for text-only data to be used in training the text encoder~\cite{pylkkonen2021fast}.

While having separate speech and text modules in RNN-Ts has its benefits, it also makes the model vulnerable to strong biases from the language model. Driven by strong textual priors, the representation from the text encoder could be very biased towards an output unit that is eventually adopted by the joint network but does not have any acoustic correlates in the speech input. Such outputs could be considered hallucinations that arise due to the overconfidence of the language model in the RNN-T~\cite{adaptLM}. This problem is more severe when the RNN-T is used to decode out-of-domain utterances. Apart from the more egregious hallucination errors, we also find that language model biases in RNN-Ts lead to word-boundary errors. For example, ``villeroy took" is mispredicted by an RNN-T baseline as ``villar I took".
\looseness=-1

Hallucinated outputs have been studied a lot more in the context of neural machine translation where the decoder language model hallucinates content that is not aligned to the source sentence~\cite{Ji_2022}. In contrast, the problem of hallucination in RNN-Ts, that stems from its very design, has been far less studied and demands more attention.

In this work, we propose \sysname\ as a fix for the problem of hallucinations in RNN-Ts. \sysname\ aims to make the textual representations more acoustically grounded by \textit{looking ahead} into the future within the speech signal. To achieve such a lookahead without interfering with the RNN-T's online decoding capabilities, we extract a limited number of lookahead output tokens for each frame of the input speech using only the acoustic encoder and further use these extracted tokens to modify the textual representation. This technique yields significant reductions in word error rates (WERs) on the established Librispeech benchmark and a variety of out-of-domain evaluation sets. We also show that beyond improving WERs, \sysname\ results in predictions that are more acoustically faithful to the speech input. 
For example, the reference ``la valliere" is misrecognized as ``the valet" by an RNN-T baseline, while an RNN-T baseline with \sysname\ predicts ``lavalier".
\looseness=-1

\textbf{Contributions}: Thus, overall the contributions of this paper are as follows: (1) We highlight the problem of hallucination in SOTA online ASR models and attribute it to the speech independent encoding of text representations.
(2) We propose a fix based on enriching text representations with a lookahead of future tokens extracted from the audio. (3) We present a simple extension of the RNN-T architecture called \sysname\  with very modest computational overheads. (4) We present an evaluation on three benchmarks on various settings of model sizes and show that our proposal improves WER and reduces hallucination significantly.

\section{Background: RNN Transducer}
\label{sec:rnnt}
Let the input audio be denoted by $\mathbf{x} = \{x_1, x_2, x_3, ... x_T \}$ where 
each $x_t$ represents the acoustic features at time $t$. Let the corresponding text transcript be $\mathbf{y} = \{y_1, y_2, y_3 ... y_U \}$ where 
 $y_u \in \mathcal{V}$ denotes the $u^{th}$ output token drawn from a vocabulary $\mathcal{V}$. 
One of the most distinguishable features of an RNN-T is the presence of two separate encoders for text and acoustic signals respectively. The acoustic encoder (AE) 
takes as input $\mathbf{x} = \{x_1, x_2, x_3, ... x_T \}$ and generates the acoustic encoded representation $\mathbf{h} = \{h_1, h_2, h_3, ... h_T \}$. The text or language encoder (LE) 
generates representation $g_u$ appropriate for the next output token as a function of previous tokens $\vy_{<u}=y_1\ldots,y_{u-1}$
\begin{align*}
    g_u = \text{LE}(\vy_{<u})~~~~~
h_t = \text{AE}(\mathbf{x},t)
\end{align*}
A \emph{Joint Network} (JN) combines the two encodings for each $t \in [1\ldots T]$ and each $u \in [1,\ldots, U]$ to generate a lattice 
$S$. Each cell of the lattice represents a state $s(t,u)$, from which we generate a probability distribution of a token belonging to vocabulary $\mathcal{V}$
    \begin{align}
    \label{eq:jn}
    P(y_u^t | x_t , \mathbf{y}_{<u}) &= \text{softmax}\{ \text{JN}(h_t \oplus g_u) \}
\end{align}
where $P(y_u^t)$ is the probability of emitting $y_u$ at time $t$.  The vocabulary includes a special blank token $\blankT$. 

Using this probability lattice, the RNN-T estimates the conditional distribution $P(\vy|\mathbf{x})$ by marginalizing over all possible monotonic alignments of acoustic frame $t$ with output token position $u$ on the lattice.
This 
can be computed in polynomial time using a DP-sum method~\cite{Graves2012}. During inference, high-probability outputs are generated using beam-search.

\begin{figure}[!]
    \centering
    \hspace*{-3.5cm}{
    \resizebox{1.45\columnwidth}{!}{
    \usetikzlibrary{positioning,arrows,shapes}
\tikzstyle{arrow}=[draw, -latex]
\begin{tikzpicture}

\node[draw, rectangle, minimum width = 3cm, minimum height = 1.25cm,label=above:{$g_u$}] (LE) {Text Encoder};
\node[draw, rectangle, minimum width = 3cm, minimum height = 1.25cm, right=0.5cm of LE, label=above:{$\mathbf{h} = \{h_1, h_2, h_3, \, \dots \, h_T\} $}] (AE) {Speech Encoder};

\node[below= 1cm of LE](input_y){$\{y_1, y_2, y_3,\,  \dots \, y_{u-1} \}$};

\node[below= 1cm of AE](input_x){$\mathbf{x}=\{x_1, x_2, x_3,\, \dots \, x_T \}$};

\node[draw, rectangle, minimum width =3cm, minimum height=1cm, fill=black!20!white, xshift=1.75cm, yshift=3cm](JN) {Joint Network};

\node[draw, dashed, rectangle, minimum width=1cm, minimum height=0.5cm,fill=black!10!white,above=1cm of LE,xshift=-0.0cm ](CN){FFN};

\node[draw,circle,right=1.08cm of CN](plus){};
\draw[-] (plus.south) -- (plus.north);
\draw[-] (plus.west) -- (plus.east);

\node[rectangle, minimum width=1cm, minimum height=1cm , above=1.15cm of JN,xshift=1.25cm] (L_iam) {$P_\text{iam}$};
\node[rectangle, minimum width=1cm, minimum height=1cm , above=1.15cm of JN] (L_rnnt) {$P_\text{rnnt}$};


\node[below= 1.4cm of LE,text=olive,label={}](ex_input_y){\small $\mathbf{y}_{<u}=\{\text{a},\text{\_},\text{fo},\text{r},\text{\_}\}$};
\node[right=-0.20cm of JN,yshift=1.15cm,text=olive,label={}] (ex_iam){\small $\begin{aligned}&\{\hat{y}_{t},\ldots,\hat{y}_{T}\} = \\ & \{\text{aa},\epsilon,\epsilon,\text{\_p},\epsilon,\text{l},\text{sa} \ldots\}\end{aligned}$};


\node [left=-1.15cm of JN, yshift=1.85cm,text=olive,label={}](ex_la){\small $\tilde{\mathbf{y}}_t^w=\{\text{aa},\text{\_p},\text{l}\}$};
\draw[->] (input_y.north) -- (LE.south);
\draw[->] (input_x.north) -- (AE.south);
\draw[->,blue] ([xshift=-0.50cm,yshift=0.5cm]AE.north) -- ([xshift=1.25cm]JN.south);
\draw[->,blue] ([xshift=1.25cm]JN.north) -- ([yshift=0.25cm]L_iam.south) node[midway, right]{}; 
\draw[->,red] ([xshift=-0.0cm,yshift=0.5cm]LE.north) -- (CN.south);
\draw[->,red] (CN.east) -- (plus.west) node[midway, above]{$g_{t,u}$};
\draw[->,blue] ([xshift=-0.5cm,yshift=1.25cm]AE.north) -- (plus.east) node[midway, sloped, above]{$h_t$};
\draw[->,blue] ([xshift=1.25cm,yshift=0.75cm]JN.north) -| (CN.north) node[pos=0.45,  above]{$\tilde{\mathbf{y}}^w_t$};
\draw[->,line width=0.25mm, green] (plus.north) -- (JN.south);
\draw[->,line width = 0.25mm, green] (JN.north) -- ([yshift=0.25cm]L_rnnt.south);

\end{tikzpicture}
    }}
\caption{Architecture of $\sysname$. We first calculate $P_\text{iam}$ which is used to calculate $P_\text{rnnt}$. An illustrative example with $w=3$\ is provided in \textcolor{olive}{olive}. $\hat{y}_{t}$ is the $\argmax$ output associated with frame $h_t$.}
\label{fig:la}
\end{figure}
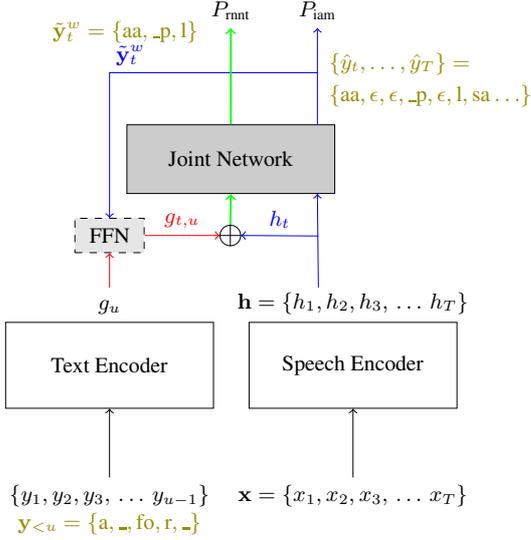

\subsection{Related Work}
The inability of RNN-T models to generalize to out-of-domain utterances has been previously well-documented~\cite{https://doi.org/10.48550/arxiv.2005.03271}. Previous approaches have adapted RNN-T models to new domains via shallow fusion with an external LM~\cite{gulcehre2015using,50651,ravi2020improving,le2020deep,zhao2019shallow} and discounting RNN-T scores using an external LM~\cite{mcdermott2020density} or an implicit LM derived from the RNN-T~\cite{variani2020hybrid,meng2020internal,adaptLM}. Apart from these inference-time techniques, there have also been other approaches that required training-time interventions such as subword regularization~\cite{le2020deep}, augmenting pronunciation dictionaries~\cite{zhao2019shallow,kitaoka2021dynamic} and text-only adaptation of the RNN-T's language encoder~\cite{pylkkonen2021fast,fnt-https://doi.org/10.48550/arxiv.2110.01500,mittal2023insitu}.

Apart from adaptation-based techniques, prior work has also explored many approaches that aim at improving interactions between the acoustic encoder (AE) and the language encoder (LE) of the RNN-T. These approaches have focused on more complex mechanisms to combine the AE and LE outputs rather than simply adding the respective representations~\cite{saon2021mult,zhang2022improving}, using quantization in the LE~\cite{VQLEShi2022} and fixing overconfident LE predictions in the RNN-T lattice~\cite{novak22_interspeech}. Similar in spirit to our technique are non-causal models ~\cite{Transformer-Transducer,emformer,Chen2020DevelopingRS,Han2020ContextNetIC,strimel2023lookahead} that refer to future information via a self-attention or convolution mechanism. However, all these methods peek into the future at the level of AE and thus with increased number of layers, the effective window size to the future increases which is not desirable in streaming applications. 

\begin{table*}[t]
\centering
\resizebox{0.975\textwidth}{!}{%
\begin{tabular}{|cc|c|c|c|c|c|c|c|}
\hline
\multicolumn{2}{|c|}{\backslashbox{Split}{Accent}}                           & US               & Eng          & Can            & Ind            & Scot         & Phil        & HK           \\ \hline
\multicolumn{2}{|c|}{Train} &
  55.3\ /\ 42023\ /\ 517 &
  \multicolumn{1}{l|}{45.0\ /\ 34615\ /\ 1856} &
  \multicolumn{1}{l|}{35.6\ /\ 24814\ /\ 575} &
  \multicolumn{1}{l|}{51.2\ /\ 34890\ /\ 982} &
  \multicolumn{1}{l|}{13.8\ /\ 8671\ /\ 85} &
  \multicolumn{1}{l|}{2.8\ /\ 2035\ /\ 47} &
  NA \\ \hline
\multicolumn{2}{|c|}{Dev}                             & 4.3\ /\ 3089\ /\ 78      & 4.2\ /\ 3465\ /\ 62  & 4.0\ /\ 3033\ /\ 68    & 5.8\ /\ 3673\ /\ 50    & 2.6\ /\ 2004\ /\ 33  & 1.5\ /\ 1009\ /\ 39 & NA           \\ \hline
\multicolumn{2}{|c|}{Test}                            & 8.0\ /\ 5363\ /\ 93     & 6.4\ /\ 5167\ /\ 71  & 6.6\ /\ 5008\ /\ 140   & 7.8\ /\ 5094\ /\ 67    & 2.8\ /\ 2001\ /\ 37  & 1.6\ /\ 1114\ /\ 22 & 3.5\ /\ 2750\ /\ 111 \\ \hline
\end{tabular}%
}

\caption{\centering {Statistics of the \textsc{MCV} dataset. Each cell lists duration (in hrs)/\# utterances/\# of speakers in the subset.}}
\label{tab:mcv_data}
\end{table*}
\begin{table*}[t]
\centering{
\resizebox{0.975\textwidth}{!}{
\small
\begin{tabular}{|l|l|llll|lllllll|}
\hline
\multirow{2}{*}{Dataset} &
  \multirow{2}{*}{Model} &
  \multicolumn{4}{c|}{Librispeech-Test} &
  \multicolumn{7}{c|}{MCV-Test} \\ \cline{3-13} 
 &
   &
  dev-clean &
  test-clean &
  dev-other &
  test-other &
  US &
  Eng &
  Can &
  Ind &
  Scot &
  Phil &
  HK \\ \hline
\multirow{2}{*}{\textsc{L100+P}} &
  Baseline &
  7.1 &
  7.6 &
  20.5 &
  20.8 &
  46.5 &
  35.6 &
  27.0 &
  69.2 &
  40.1 &
  46.3 &
  48.1 \\
 &
  \sysname &
  \textbf{6.4} &
  \textbf{6.6} &
  \textbf{18.8} &
  \textbf{19.1} &
  \textbf{44.9} &
  \textbf{33.2} &
  \textbf{24.8} &
  \textbf{68.6} &
  \textbf{37.3} &
  \textbf{42.6} &
  \textbf{45.7} \\ \hline
\multirow{2}{*}{\textsc{L360+P}} &
  Baseline &
  4.2 &
  4.5 &
  13.5 &
  13.1 &
  38.1 &
  23.6 &
  19.5 &
  56.9 &
  30.2 &
  34.4 &
  39.1 \\
 &
  \sysname &
  \textbf{4.0} &
  \textbf{4.3} &
  \textbf{12.8} &
  \textbf{12.5} &
  \textbf{36.9} &
  \textbf{22.5} &
  \textbf{19.2} &
  \textbf{56.6} &
  \textbf{28.9} &
  \textbf{33.2} &
  \textbf{37.5} \\ \hline
\multirow{2}{*}{\textsc{L960}} &
  Baseline &
  \textbf{3.1} &
  \textbf{3.2} &
  8.6 &
  8.5 &
  29.2 &
  15.0 &
  16.1 &
  35.5 &
  23.1 &
  23.9 &
  29.7 \\
 &
  \sysname &
  3.2 &
  3.3 &
  \textbf{8.3} &
  \textbf{8.4} &
  \textbf{28.8} &
  \textbf{14.6} &
  \textbf{15.9} &
  \textbf{35.2} &
  \textbf{23.0} &
  \textbf{23.8} &
  \textbf{29.0} \\ \hline
  \multirow{2}{*}{\textsc{MCV}} &
  Baseline &
  18.9 &
  18.8 &
  31.4 &
  31.7 &
  27.9 &
  18.1 &
  19.1 &
  27.4 &
  25.8 &
  27.2 &
  31.9 \\
 &
  \sysname &
  \textbf{16.1} &
  \textbf{15.8} &
  \textbf{27.6} &
  \textbf{28.2} &
  \textbf{23.8} &
  \textbf{14.3} &
  \textbf{15.3} &
  \textbf{24.2} &
  \textbf{21.9} &
  \textbf{23.9} &
  \textbf{27.8} \\ \hline
\end{tabular}%
}
\caption{Main results of baseline vs \sysname.}
\label{tab:main_table}
}
\end{table*}
\begin{table}[hb!]
\resizebox{\columnwidth}{!}{%
\begin{tabular}{|l|cccccccccccccccccccccccccccc|}\hline
\multirow{2}{*}{} & \multicolumn{28}{c|}{Libri-test}                                                                                                                                                                                                                                        \\ \cline{2-29} 
                       & \multicolumn{7}{c|}{dev-clean}                                   & \multicolumn{7}{c|}{test-clean}                                  & \multicolumn{7}{c|}{dev-other}                                  & \multicolumn{7}{c|}{test-other}                                 \\ \hline
Baseline               & \multicolumn{7}{c|}{\textbf{20.0}}                              & \multicolumn{7}{c|}{21.7}                                       & \multicolumn{7}{c|}{48.3}                                      & \multicolumn{7}{c|}{51.6}                                      \\
\sysname               & \multicolumn{7}{c|}{20.3}                                       & \multicolumn{7}{c|}{\textbf{20.1}}                              & \multicolumn{7}{c|}{\textbf{47.2}}                             & \multicolumn{7}{c|}{\textbf{50.7}}                             \\ \hline
\multirow{2}{*}{} & \multicolumn{28}{c|}{MCV-test}                                                                                                                                                                                                                                          \\ \cline{2-29} 
                       & \multicolumn{4}{c|}{US}             & \multicolumn{4}{l|}{Eng}            & \multicolumn{4}{l|}{Can}            & \multicolumn{4}{l|}{Ind}            & \multicolumn{4}{c|}{Scot}           & \multicolumn{4}{c|}{Phil}           & \multicolumn{4}{c|}{HK}             \\ \cline{1-29} 
Baseline               & \multicolumn{4}{c|}{61.1}          & \multicolumn{4}{l|}{63.4}          & \multicolumn{4}{l|}{46.2}          & \multicolumn{4}{l|}{85.2}          & \multicolumn{4}{c|}{60.7}          & \multicolumn{4}{c|}{61.2}          & \multicolumn{4}{c|}{70.7}          \\
\sysname               & \multicolumn{4}{c|}{\textbf{58.7}} & \multicolumn{4}{l|}{\textbf{59.9}} & \multicolumn{4}{l|}{\textbf{41.5}} & \multicolumn{4}{l|}{\textbf{84.4}} & \multicolumn{4}{c|}{\textbf{58.0}} & \multicolumn{4}{c|}{\textbf{58.3}} & \multicolumn{4}{c|}{\textbf{67.7}} \\ \hline
\end{tabular}%
}
\caption{Rare-WER comparison of baseline vs \sysname\ on \textsc{L100+P}.}
\label{tab:rare_table}
\end{table}

\section{Our Approach}
\label{sec:lookahead}
In the RNN-T architecture the text and the audio are encoded independently and combined  using a thin joint layer.  The text encoding $g_u$ captures the next token state conditioned only on tokens generated before it.  If the language model priors are strong, it is possible for the text encoding $g_u$ to strongly prefer a word unit, and to bias the softmax to generate that unit, even if that has little agreement with the acoustic input.  This often manifests as hallucinated words that are fluent as per the language model, but which have no support with the acoustic input.  In Table~\ref{tab:anecdotes2.tex}, we present examples of such hallucinations. We observe that rare words, e.g., word ``MAGATAMA" in the first example, gets recognized as ``BY THE TOWN" which has little acoustic overlap with the utterance.

%

Next we present our approach called \sysname\ of fixing this mismatch. 
We identify that hallucinations are generated by representations $g_u$ which have been computed independent of the acoustic input. 
%
%
Our key insight is to improve the LE representation $g_u$ by looking ahead a few tokens into the future from the acoustic input.  However, implementing this insight, without adversely impacting the efficient online operation of the RNN-T model is non-trivial.  We perform this modification in two steps.  First, we extract from the acoustic encoder, a lookahead of $k$ tokens after each frame $t$ of the acoustic input.  In Section~\ref{sec:futureToks} we present how we do that.  Second, we modify $g_u$ with the extracted future tokens to get now a time-aware, acoustically informed encoding $g_{u,t}$.  Section~\ref{sec:gu} presents the details of this step. Overall modified architecture of our approach appears in Figure~\ref{fig:la}.
\begin{table*}[t]
\centering
\resizebox{\textwidth}{!}{%
\ttfamily
\bfseries
\begin{tabular}{|l|l|l|}
\hline
\multicolumn{1}{|c}{Reference}                                & \multicolumn{1}{|c}{Baseline}                         & \multicolumn{1}{|c|}{ \sysname}                         \\ \hline
shaped like commas or \text{\textcolor{blue}{MAGATAMA}}     & sheep like commas or \text{\textcolor{blue}{BY THE TOWN}} & sheep like commas or \text{\textcolor{blue}{MAGATINA}}      \\
confession is \text{\textcolor{blue}{GOOD FOR}} the soul &
  the fashion is \text{\textcolor{blue}{GIVEN FORWARD}} the soul &
  confession is \text{\textcolor{blue}{GOOD FOR}} the soul \\
for \text{\textcolor{blue}{ALL WHO SIN ALL WHO SUFFER}} you 
& for \text{\textcolor{blue}{OLDER SIN OLD RISK}} you       
& for \text{\textcolor{blue}{ALL WHOOSIN OLD WHO SUFFER}} you \\   
\text{\textcolor{blue}{LIFE WAS}} not \text{\textcolor{blue}{EASY}}                  & by \text{\textcolor{blue}{OR}} not \text{\textcolor{blue}{EVEN}}                   & \text{\textcolor{blue}{LIFE OR}} not \text{\textcolor{blue}{EASY}}                   \\
do \text{\textcolor{blue}{ANDROIDS}} dream of &
do \text{\textcolor{blue}{ENJOYS}} dream of &
  do \text{\textcolor{blue}{ANDREWS}} dream of \\
\text{\textcolor{blue}{LA VALLIERE}} is quite a poetess said &
  \text{\textcolor{blue}{THE VALET}} is quite poetic said &
  \text{\textcolor{blue}{LAVALIER}} is quite a poeticus said \\

\hline
\end{tabular}%
}
\caption{Anecdotes comparing \text{Baseline} with \sysname. Baseline hallucination errors are highlighted in \text{\textcolor{blue}{blue}}.}
\label{tab:anecdotes2.tex}
\end{table*}


\subsection{Extraction of Lookahead tokens from the Acoustic Encoder}
\label{sec:futureToks}
 
We use the notion of \emph{implicit acoustic model} (IAM) and generate a probability distribution over vocabulary tokens
based on only the acoustic signals $h_t$
\begin{equation}
\label{eq:iam}
    \IAM(y|h_t) = \text{softmax}\{ \text{JN}(h_t \oplus \mathbf{0}) \}
\end{equation}
This distribution is supervised using gold transcripts by marginalizing over all non-blank tokens, much like the regular RNN-T loss.

Next, for each frame of $\vx$ we get the most likely output based on acoustic signal $h_t$.
\begin{equation}
    \hat{y}_{t} = \argmax_y \IAM (y|h_t)~~~\forall t
    \label{eq:IAM}
\end{equation}
We do this for all input frames.  On this set of $T$ tokens, at each $t$, we find the first $w$ non-blank tokens after $t$ and call it $\tilde{\vy}_t^w$.  We choose small values of $w$ (e.g., 2 or 3) to ensure that we still inherit the streaming capabilities of the underlying RNN-T model. The $\tilde{\vy}_t^w$ provides a look ahead into the future tokens after each $t$ based purely on the acoustic input.  Figure~\ref{fig:la} shows an example where for $t=1,w=3$ we extract a  $\tilde{\vy}_t^w$ comprising of the first three non-blank outputs from the acoustic model.

\subsection{Modifying Text Encoding with Lookahead Tokens}
\label{sec:gu}
With the presence of the above future context, we condition the text encoding $g_u$ at each $t$ using a simple FFN $F$ as
\begin{equation}
    \hat{g}_{t,u} = F(g_u, \tilde{\vy}_t^w)
\end{equation}
The remainder of the RNN-T pipeline stays the same and we pass along the new LE representation $\hat{g}_{t,u}$ to the JN for further processing
\begin{equation}
        P_\text{LA}(y_u^t | \vx, t, \mathbf{y}_{<u}) = \text{softmax}\{  \text{JN}(h_t \oplus \hat{g}_{t,u}) \}
\end{equation}

Thus, our overall training is a sum of losses over the gold transcript on marginalizations of  $P_\text{LA}$, and of $\IAM$.
\begin{table*}[t]
\centering
\resizebox{\textwidth}{!}{%
\begin{tabular}{|l|l|llll|lllllll|}
\hline
 &
  Model &
  dev\_clean &
  test\_clean &
  dev\_other &
  test\_other &
  us &
  can &
  eng &
  ind &
  scot &
  phil &
  hk  \\ \hline
\multirow{2}{*}{PER} &
  Baseline &
  2.52 &
  2.73 &
  9.90 &
  9.81 &
  24.46 &
  11.74 &
  17.34 &
  39.33 &
  19.55 &
  21.35 &
  23.96 \\
 &
  \sysname &
  \textbf{2.28} &
  \textbf{2.35} &
  \textbf{8.99} &
  \textbf{8.98} &
  \textbf{23.40} &
  \textbf{10.42} &
  \textbf{15.80} &
  \textbf{38.18} &
  \textbf{17.55} &
  \textbf{19.20} &
  \textbf{22.41} \\ \hline
   \multirow{2}{*}{WFED} &
  Baseline &
 8.31 &
  8.95 &
  31.53 &
  31.25 &
  77.95 &
  39.07 &
  52.89 &
  118.89 &
  63.40 &
  64.02 &
  73.55 \\
 &
  \sysname &
  \textbf{7.57} &
  \textbf{8.00} &
  \textbf{28.84} &
  \textbf{29.13} &
  \textbf{74.47} &
  \textbf{34.20} &
  \textbf{48.23} &
  \textbf{115.30} &
  \textbf{57.12} &
  \textbf{58.94} &
  \textbf{69.04}
   \\
  \hline
  \multirow{2}{*}{DER} &
  Baseline &
  1.13 &
  1.22 &
  4.37 &
  4.34 &
  11.48 &
  5.52 &
  7.23 &
  17.50 &
  9.00 &
  8.98 &
  10.23 \\
 & \sysname & \textbf{1.03} & \textbf{1.10} & \textbf{4.01} & \textbf{4.03} & \textbf{11.01} & \textbf{4.83} & \textbf{6.58} & \textbf{16.82} & \textbf{8.10} & \textbf{8.27} & \textbf{9.57}  \\
  \hline
 
\end{tabular}%
}
\caption{Comparing acoustically-aware error metrics between \text{Baseline} and \sysname\ trained using \textsc{L100+P}.}
\label{tab:pers}
\end{table*}
\section{Experiments and Results}
\label{sec:results}

In this section, we present various results comparing our method to an existing SoTA baseline RNN-T ASR system.
\subsection{Datasets}
We use the popular Librispeech benchmark~\cite{7178964} in three different well-known settings of varying training durations. We also use the Mozilla Common Voice dataset to derive accented out-of-domain speech samples. These are detailed below:
\begin{enumerate}
    \item \textsc{L100}: This is the well-known Librispeech-100 dataset which is a 100 hour subset of the Librispeech corpus\cite{7178964}.
    \item \textsc{L360}: This is a 360 hour subset of the Librispeech corpus.
    \item \textsc{L960}: This is the complete 960-hour Librispeech corpus.
    \item \textsc{MCV}: This is a dataset obtained from the Mozilla Common Voice \cite{commonvoice:2020} Version 7.0 corpus. Using this dataset, we create a 200hr training subset consisting of seven English accents of which one is a test only accent. The accent-wise composition of the \textsc{MCV} dataset can be found in Table~\ref{tab:mcv_data}.
\end{enumerate}
\textsc{X+P} refers to systems with speed-perturbation that augments the dataset \textsc{X} with time-warped copies using factors $0.9, 1.0, 1.1$ respectively. In Table~\ref{tab:main_table}, we indicate wherever this augmentation is used. We test the above models both on in-domain and out-of-domain test splits, where the latter is obtained by changing the accent from the MCV corpus.  


\subsection{Implementation Details}
We use the ESPNet toolkit \cite{watanabe2018espnet} to implement our model and perform all our experiments. Our base architecture is an RNN-T model with conformers~\cite{gulati2020conformer} as AE and LSTM~\cite{lstm97} as LE. The joint network adds the two encodings and applies a $\tanh$ non-linearity. We experiment with two different models corresponding to different encoder sizes. 
\begin{itemize}
    \item \textsc{Small:} This model includes $18$ AE blocks with internal representation of size $256$ and four attention heads. A macaron-style feed forward projecting to size 1024 is used within the conformer block. The LE consists of a single $300$ dimensional LSTM cell.  Both these encoders project their output to a joint-space of $300$ dimensions before feeding it to the joint network. This model was used for performing experiments with the \textsc{L100+P} and \textsc{MCV} datasets. We use a vocabulary size of $300$ and $150$ sub-words for these experiments, respectively ($\sim30$M params).
    \item \textsc{Large:}  This model has $12$ AE blocks conformer-encoder  of size $512$ and a macaron style feed-forward of $2048$. The LE block is a $512$ unit LSTM. Both representations are projected to a joint-space of of $512$ dimensions.  This model is used for our experiments involving the \textsc{L960} and \textsc{L360+P} datasets. We use a vocabulary size of $500$ for these experiments ($\sim 80$M params). 
\end{itemize}
We train for $100$ epochs using the \emph{Noam}-optimizer \cite{vaswani2017attention,adam} with a Noam-learning rate of 5 and 25000 warm-up steps. For inference, we use ALSD~\cite{alsd-9053040} with a beam size of 30. For the purposes of \sysname\ we have used $3$ lookahead tokens. 


\subsection{Results}
Our primary set of numbers can be seen in Table~\ref{tab:main_table} which shows the capability of \sysname\ against baseline models trained on various training data sizes. Using \textsc{L100+P}, we see statistically significant improvements ($p < 0.01$ using the MAPSSWE test~\cite{gillick1989some}) in WERs across both training sets for in-domain and out-of-domain evaluation sets. The gains are reduced with using larger models (\textsc{L360+P} and \textsc{L960}) but still remain consistent, especially for the out-of-domain test sets. We further focus on rare words where the effect of LM biasing is expected to have a greater effect.   Table~\ref{tab:rare_table} shows WER  performance of baseline vs \sysname\ on rare-words on the  \textsc{L100+P} dataset. We define any word that occurs lower than $20$ times in the dataset as a rare-word. We see consistent improvement in rare-word recognition across both in-domain as well as out-of-domain datasets. Table~\ref{tab:anecdotes2.tex} shows some anecdotes indicating the effectiveness of \sysname. Note how the baseline model hallucinates words like TOWN, GIVEN, etc. that have no acoustic overlap with the spoken word.  \sysname\ either corrects them or produces a word that is more acoustically similar to the spoken word.

\begin{table}[b]
\centering
\begin{tabular}{|l|llc|}
\hline
Metric & Ref                  & Hyp                        & Dist \\ \hline
PER    & ball (\textipa{bOl}) & maul(\textipa{mOl})        & 1    \\
                        & call (\textipa{kOl}) & install (\textipa{InstOl}) & 4    \\ \hline
WFED   & ball (\textipa{bOl}) & paul (\textipa{pOl})       & 0.25 \\
                        & ball (\textipa{bOl}) & call (\textipa{kOl})       & 2.25 \\ \hline
DER    & pit (\textipa{pIt})  & meet (\textipa{mit})       & 1    \\
                        & pit (\textipa{pIt})  & beat (\textipa{bit})       & 0    \\ \hline
\end{tabular}%
\caption{Illustration of different error rate metrics}
\label{tab:per_examples}
\end{table}



\subsection{Error Analysis}
Apart from WERs, we provide further analysis in Table~\ref{tab:pers} to validate our claim that predictions from \sysname\ are indeed acoustically more similar to the ground-truth. We use the following three metrics:
\begin{itemize}
    \item Phone Error Rate (PER): We use Epitran \cite{Mortensen-et-al:2018} to transcribe word sequences to phone sequences in IPA and find the IPA-based edit distance to yield PERs, which are a better measure of acoustic discrepancy compared to WERs.
    \item Weighted Feature-based Edit Distance (WFED): With the PER metric, the distance between two dissimilar phones and two related phones are both 1.
    WFED is an edit distance metric based on Panphon~\cite{Mortensen-et-al:2016} that helps compute distances between IPA phones based on their articulatory features. 
    \item Dolgopolsky Error Rate (DER): Both PER and WFED rely on an underlying sequence of IPA phones derived from the predicted word sequences using a fixed set of rules. These IPA sequences are not very reliable, especially for the out-of-domain MCV speech samples with varying speech accents. To alleviate errors arising from incorrect IPA, we use Dolgoposky's IPA clusters \cite{dolgo} to label similar-sounding IPA phones with the same cluster ID.  
    DER is an edit distance using these new cluster IDs that only measures large deviations in phonetic realizations between a reference word and a predicted word and does not penalize small changes in the place and manner of articulation of phones (like PER).
\end{itemize}
\begin{table}[]
\centering
\resizebox{0.9\columnwidth}{!}{%
\small
\begin{tabular}{|l|cccc|}
\hline
$w$ & dev-clean    & test-clean   & dev-other     & test-other    \\ \hline
2 & 6.5          & 6.9          & 18.8          & 19.0          \\
3 & \textbf{6.4} & \textbf{6.6} & \textbf{18.8} & 19.1 \\
5 & 6.4          & 6.8          & 18.9          & \textbf{18.9}          \\ \hline
\end{tabular}%
}
\caption{Ablation on \textsc{L100+P} by varying $w$ with \sysname.}
\label{tab:ablation_window}
\end{table}
 Table~\ref{tab:per_examples} lists illustrative examples for each of the three metrics. 
 E.g., WFED imposes a cost of 2.25 if ``ball" is misrecognized as ``call" (which would have a PER of 1); DER imposes no cost for ``pit" being misrecognized as ``beat". Table~\ref{tab:pers} compares PER, WFED and DER values from the baseline and \sysname. 
 DER penalizes only for large acoustic digressions; we see consistent improvements on DER using \sysname\ with larger improvements on the MCV test utterances. We also show ablations on the window size of future tokens considered in Table~\ref{tab:ablation_window}. We observe that using a window size of $3$ yields the best results.



\section{Conclusion}

We propose a simple and effective scheme of acoustic \sysname\ for RNN-T models to safeguard against producing acoustic hallucination. From the acoustic encoder, we extract a fixed \sysname\ of tokens and use it to improve the representation from the language encoder to generate a more acoustically aligned output. We obtain significant reductions in WER and various other phonetic error rates across various settings.   


\section{Acknowledgements}
The third author gratefully acknowledges financial support from a SERB Core Research Grant, Department of Science and Technology, Govt of India on accented speech processing.
\newpage

\bibliographystyle{IEEEtran}
\bibliography{bibliography}

\end{document}